# License Plate Recognition (LPR): A Review with Experiments for Malaysia Case Study

Nuzulha Khilwani Ibrahim, Emaliana Kasmuri, Norazira A Jalil, Mohd Adili Norasikin, Sazilah Salam
Faculty of Information and Communication Technology,
Universiti Teknikal Malaysia Melaka (UTeM), Hang Tuah Jaya, 76100 Melaka, Malaysia
nuzulha | emaliana | adili | sazilah @utem.edu.my

Mohamad Riduwan Md Nawawi
Faculty of Electrical Engineering,
Universiti Teknikal Malaysia Melaka (UTeM), Hang Tuah Jaya, 76100 Melaka, Malaysia
riduwan @utem.edu.my

*Abstract*— **Most vehicle license plate recognition use neural network techniques to enhance its computing capability. The image of the vehicle license plate is captured and processed to produce a textual output for further processing. This paper reviews image processing and neural network techniques applied at different stages which are preprocessing, filtering, feature extraction, segmentation and recognition in such way to remove the noise of the image, to enhance the image quality and to expedite the computing process by converting the characters in the image into respective text. An exemplar experiment has been done in MATLAB to show the basic process of the image processing especially for license plate in Malaysia case study. An algorithm is adapted into the solution for parking management system. The solution then is implemented as proof of concept to the algorithm.**

*Keyword- image processing, preprocessing, filtering, feature extraction, segmentation, recognition, experiment*

## I. INTRODUCTION

The advanced of computer application processed more than textual data solving everyday problems. Inputs from optical device are used in domain application such as medical, security, monitoring and control and engineering. Ability for computer to process image and translate it into something meaningful has become more popular. Therefore, the technology of image processing has adopted in managing vehicle parking system, vehicle access to restricted area, traffic monitoring system and highway electronic toll collection. For this purpose, the computer needs to capture the vehicle licence plate number and process it in the computer.

A camera captures the image of vehicle license plate. The image then feed into the computer for further processing. The output of from the process is the vehicle license plate number in textual form. For a parking system, the output is used for car identification, parking payment and authorization to access into the parking space. This paper reviews the processing of vehicle license plate that uses image processing and neural network technique.

The framework for this research is adapted from previous studies [1-4] as shown in Figure 1 which includes 5 stages: (a) pre-processing, (b) filtering, (c) feature extraction, (d) segmentation and (e) character recognition. The final output of the sample experiment is to recognize the alphanumeric characters on the license plate. The structure of this paper is organized by the stages of the process.

## II. PREPROCESSING

Digital image preprocessing is an initial step to image processing improving the data image quality for more suitable for visual perception or computational processing. Preprocessing remove unwanted data and enhance the image by removing background noise, normalizing the intensity of individual image particles, image deblur and remove image reflections. Preprocessing for car license plate number uses three common subprocesses, which are geometric operation, grayscaling process and binarization process.

Many neural network techniques have been applied to these preprocessing techniques mainly to produce better image and to increase the speed of convergence of an image.

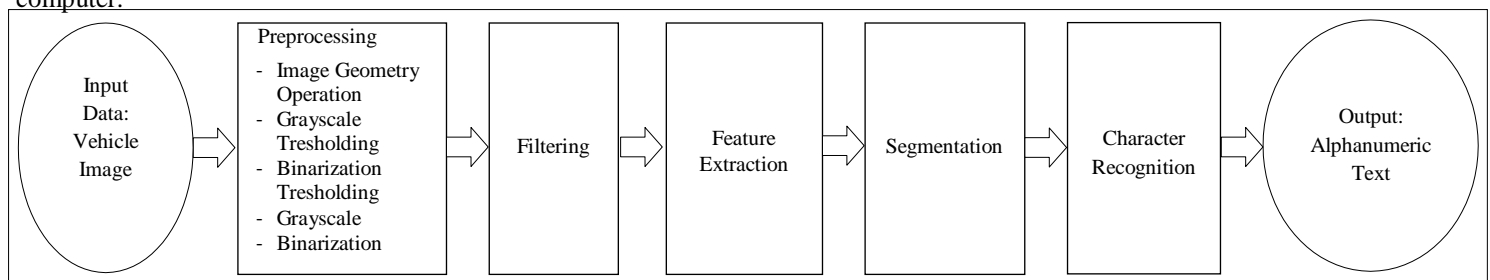

Figure 1. Research framework

The research is funded for Malaysian Technical University Network - Centre of Excellence (MTUN CoE) program with grant no MTUN/2012/ UTeM-FTMK/11 M00018.

### A. Geometric Operation

Geometric operation is a process to locate the car license plate. The purpose of this operation is to localize the car plate for faster character identification over a small region.

An improved Back Propagation network is used to overcome the weakness of convergence speed in [1]. Genetic algorithm and momentum term is introduced to the current network to increase the speed of convergence rate. The current BP network learning process is said to be easily produce error if initial weights is not set properly [1] and it is difficult to determine the number of hidden layer and hidden nodes. The improved network using BP momentum increase the speed and the accuracy to localize the car license place location. A grayscale image extracts the edge of the license plate using sobel operator [1].

Malviya and Bhirud in [2] uses iterative thresholding operation to identify license plate of a vehicle. Objects with geometric characteristics are labelled and selected. The process takes into account aspect ratio, total pixel per object, height, width and the presence of characters in the region.

For this, we propose the following algorithm, where the pseudo-code can be simplified as the following:
- to get the scale of the image for x-axis and y-axis
- to assign the new value of horizontal and vertical axis based on the scale of the x-axis and y-axis
- to get the grayscale thresholding value of the image

The input of the experiment is shown as Figure 2 while the example output can be viewed as Figure 3.

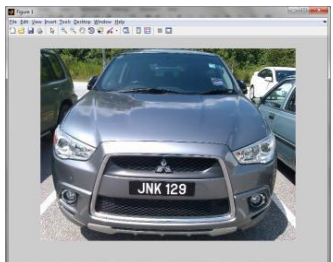

Figure 2. Input data for License Plate Image Processing

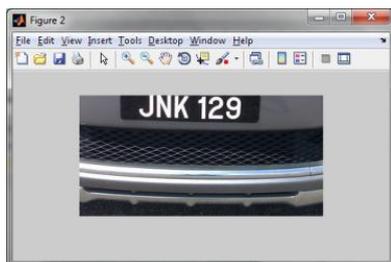

Figure 3. Image output after geometric operation process

The output from the extraction process will be used in the next stage which is grayscaling process.

### B. Grayscaling Process

Grayscaling is a process to produce a gray scale image from a multicolor image. In this process, the threshold of an image is calculated. If it is less than the threshold, the image data is recalculated to get the correct grayscale value. The purpose of thresholding is to separate the object of interest from the background. Thresholding is important to provide sufficient contrast for the image so that different level of intensity between object and the background can be differentiated for later computational processing. Different intensity determines the value of the threshold.

Grayscaling process improves the quality of the image for later computational processing. Other preprocessing techniques to improve the quality of the image including image deblurring, image enhancement, image fusion and image reconstruction.

Image fusion is a process to enhance the image with multiple combinations of images [2-3]. This process is suitable to identify the car license registration number from a moving car. The technique integrates multi resolution image and produce a composite image using inverse multiresolution transform [3]. A template of image from a grayscale is shifted to vertical and horizontal direction. The contrast frequency is calculated for each position in the template and creates a new image using thresholding procedure. Any color below the threshold is set to back (zero) and above threshold is set to white (one). The value determines the gray level resulting black and white image.

A trained feedforward neural network (FFN) with Block Recursive LS algorithm is used to process car license plate [4]. The approach is to improve the convergence rate and stabilize the robustness of the solution. The location of the car license plate is extracted using Discrete Fourier Transform (DFT). DFT identifies maximum value of horizontal and vertical edges. Prior to that tone equalization and contrast reduction is used to improve the image. These techniques are preferred because it is more robust and suitable compared to edge enhancement.

For this, we propose the following algorithm, where the pseudo-code can be simplified as the following:
- to convert into grayscale image

The pseudo-code can be translated in MATLAB such as following:
- TestImg1 =rgb2gray(TestImg1);

### C. Binarization Process

Binarization is a process of converting grayscale image into black and white image or "0" and "1". Previously, the gray scale image consists of different level of gray values; from 0 to 255. To improve the quality and extract some information from the image, the image needs to be process a few times and thus make the binary image more useful. Gray threshold value of an image is required in the binarization process as it is important to determine whether the pixels that having gray values will be converted to black or white.

For this, we propose the following algorithm, where the pseudo-code can be simplified as the following:
- to convert into black and white image

The pseudo-code can be translated in MATLAB such as following:

- ImgBW =im2bw(TestImg1 ,threshold

The example output can be viewed as Figure 4.

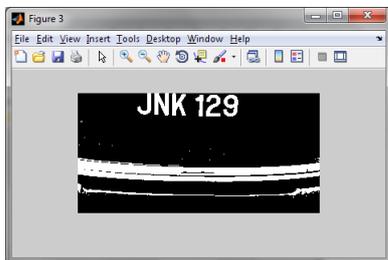

Figure 4. Image output after binarization process

The output from the extraction process will be used in the next stage of the processing in this framework which is filtering.

### III. FILTERING

To enhance the quality of processing image, filtering is required to solve contrast enhancement, noise suppression, blurry issue and data reduction. It is reported that most of preprocessing activities conducted in image restoration apply Neural Network approach [5].

Rectangles" filtering implemented on the real plate number involves convolution matrix, binarization filter with vertical and horizontal projection able to enhance the image quality and eliminates unwanted pieces on the plate. It is also recognize the number of rows and symbols in the plate number [6].

In [7], a simple filter is designed by implementing intensity variance and edge density to overcome illumination issue, distance changed and complex background. It is proposed that this approach convenient for real-time application.

The quality and selection of parameters on the camera extremely contributes the desired preprocessing image quality [8].

The example output can be viewed as Figure 5. The output from the extraction process will be used in the next stage which is image segmentation.

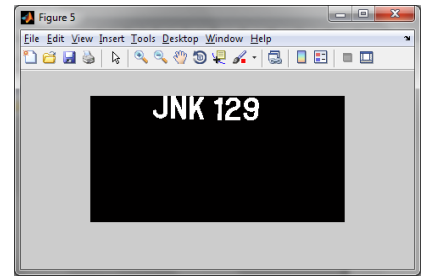

Figure 5. Image output after filtering process

### IV. FEATURE EXTRACTION

Features extraction is the part of measuring those relevant features to be used in recognition process. Selection of the right features is important in order to obtain best results in license plate recognition study. Colour features are very good potential for object detection. However the parameter such as colour of car, illumination condition and the quality of imaging system has been limited its practice [9]. According to [10], colour features have been studied by [11] and [12] but from the study, this feature not robust enough to various environments. However, there are many types of features that can aid license plate recognition such as aspect ratio, texture, edge density, and size of region [10]. In order to achieve better detection rate in license plate recognition, researchers in [10] and [13] had suggested a combination of features. For instance, a promising result for combination of colour and edge has been reported in [14]. Moreover, [9] has reported that the use of simple geometrical features such as shape, aspect ratio, and size are enough to find genuine license plate. However the researchers face problem such as clutter parts in the image and overcome it with edge density. Edge features of the car image are very important, and edge density can be used to successfully detect a number plate location due to the characteristics of the number plate [9]. The edge density features had been used in [9, 10, 13] because the density of vertical edges at the license plate area is considerably higher than its neighbourhood. In addition, this feature is more reliable and able to reduce processing time. Little computational time is one of important element in recognition especially in real-time detection. However, there is always trade-off between the number of features used in the system and the computational time [9, 13].

For this, we propose the following algorithm, where the pseudo-code can be simplified as the following:
- To compare the vertical and horizontal histogram in getting the required features.
- to extract the meaningful image based on the features selected

Then, the horizontal and vertical histograms are combined to get the matching region of a license plate is kept as candidate region or also known as meaningful image. The example output can be viewed as Figure 6 and Figure 7.

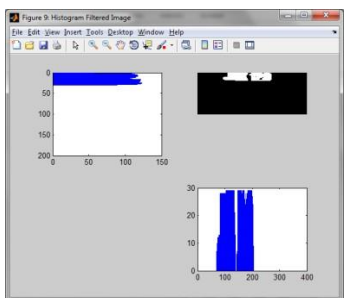

Figure 6. Image output after feature selection process

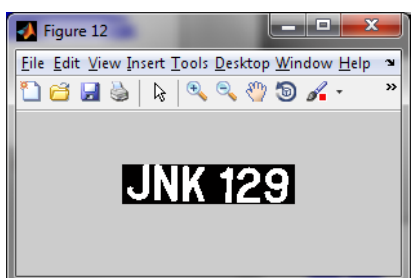

Figure 7. Image output after feature extraction process

The output from the extraction process will be used in the next stage which is image segmentation.

## V. IMAGE SEGMENTATION

One of the most popular topics in image processing study is image segmentation. The segmentation process becomes important to the processing of the image to find the meaningful information where it comes from the meaningful regions which represent higher level of data. The analysis of image requires large amount of low level of data which is in pixel to be extracted into meaningful information.

Higher-level object properties can be incorporated into segmentation process, after completing certain preliminary segmentation process. Examples of higher-level properties are as follow:
i. shape, or
ii. colour features

Then, it comes to the goal of segmentation which is to find regions that represent meaningful parts of objects. In segmentation, the image will be divided into regions based on the interest of the study.

Image segmentation methods will look for objects that either have some measure of homogeneity (within themselves), or contrast (with the objects on their border). Most image segmentation algorithms can be divided as the following:
i. modifications,
ii. extensions, or
iii. combination of these 2 basic concepts

Classically, Umbaugh in [15] divide image segmentation techniques into three (3) which are:
i. Region growing and shrinking: subset of clustering
ii. Clustering methods
iii. Boundary detection: extensions of the edge detection techniques

At the same point, Haralick and Shapiro [16] categorized image segmentation techniques into six (6) which are:
i. Measurement space guided spatial clustering
ii. Single linkage region growing schemes
iii. Hybrid linkage region growing schemes
iv. Centroid linkage region growing schemes
v. Spatial clustering schemes
vi. Split and merge schemes

Clustering is one of the segmentation technique as Haralick and Shapiro [16] differentiated clustering and segmentation such as follow:
i. In clustering: the grouping is done in measurement space
ii. In segmentation: the grouping is done in the spatial domain of the image

Clustering techniques can be used to any domain, eg: any N-dimensional color or feature space, including spatial domain"s coordinates. This technique segments the image by placing similar elements into groups, or clusters, based on some similarity measure. Clustering is differ from region growing and shrinking methods, where the mathematical space used for clustering. The details of each methods in segmentation are explained in the next sections.

A. *Thresholding*

Thresholding is one of the simplest and most popular method in image segmentation. Two common types of thresholding are outlined as follow:
i. Local thresholding is referred when an image is partitioned into subregions, and each subregion carry different value of threshold. Local threshold method also called as adaptive thresholding schemes [17-19].
ii. Global thresholding is referring to assigning only one threshold value to the entire image.

Thresholding techniques also can be categorized into two levels:
i. Bilevel thresholding: the image is two (2) regions which are object (black) and background (white).
ii. Multithresholding: the image is composed of few objects with different surface characteristics thus need multiple value of threshold.

Thresholding also can be analyzed as classification problem, such that classifiying bilevel segmentation of an image into object and background. Among the most common methods found for thresholding in image segmentation are listed as the following:
i. maximum entropy method [20-22]

    ii.    Otsu's method (maximum variance) [23-26]
    iii.   k-means clustering [27-35]

### B. Edge Detection

There will be edge and line detection in segmentation to divide regions into meaningful information. Edge detection techniques: Line detection/ line finding = Hough transform [37]. Hough transform is designed specifically to find lines. A line is a collection of edge points (that are adjacent and have the same direction). The Hough algorithm will take a collection of few edge points.

Edge detection techniques [38-53] have been used as the base of another segmentation technique. Basically, edge detection is also an independent process in image processing. Edge detection, or sometimes it is called as edge finding is also closely related to region detection. We need to find the region boundaries first before we can proceed to segment an object from an image. This is because the edges identified by edge detection are frequently disconnected. It means that we have to find the boundaries in order to get the edges.

In segmentation, line detection is done to divide regions into meaningful information. One of line detection technique is Hough transform. Hough transform is designed specifically to detect lines. A line is a collection of edge points (that are adjacent and have the same direction). The Hough algorithm will take a collection of few edge points.

### C. Region-based image segmentation

This technique attempt to classify a particular image into several regions or classes according to the common properties of the image. There are few properties considered for this process which are pattern and texture, intensity values and spectral profiles of the image. In this method, we want to group the regions so that each of the pixels in the region will have similar value of the properties. There are many real applications used this method such as remote sensing, 2D and 3D images [54-55] while there are various models and algorithms used for this technique such as Markov Random Field Model [56-60] and Mumford-Shah Algorithm [61-64].

### D. Compression-based methods

In this method, segmentation will be done in a way the image will be compressed based on the similarity of the patterns of textures or boundary shape of the image. This method aims to minimize the length of the data where the optimal segmentation can be achieved. There are few ways on how to calculate the coding length of the data such as Huffman coding or MDL (Minimum Description Length) principle [65-66], where they can be found in previous studies [67-70].

### E. Histogram-based methods

Histogram-based method [71-76] is one of the frequently used for image segmentation techniques. In this method, we will produce a vertical and a horizontal histogram accordingly. This process is to get a group of pixels in vertical and horizontal regions where they will lead to distinguishing the gray levels of the image.

In common, an image will have two regions: background and object. Normally, the background is assigned as one gray level whiles the object (or also called as subject) is another gray level. Usually, background will secure the largest part of the image so the gray level of it will have larger peak in the histogram compared to the object of the image.

### F. Region-growing methods

Region Growing and Shrinking [77-99] technique use row and column (r,c) based image domain. It can be considered as subset of clustering methods, but limited to spatial domain. The methods can be:
- Local : operating on small neighbourhoods, or
- Global : operating on the entire image, or
- Combination of both

### G. Split-and-merge methods

There is an alternative for segmentation method called split and merge [100-108]. Split and merge is also called as quadtree segmentation where it based on quadtree partition. The data structure used in split and merge is called quadtree where a tree which has nodes and each node can have four children. It divides regions that do not pass a homogeneity test, and combines regions that pass the homogeneity test.

For this, we propose the following algorithm, where the pseudo-code can be simplified as the following:
- to get width of y-axis of the image to divide into subregion
- to get width of y-axis of the image to divide into subregion
- to divide into subregion
- to remove blank space
- to get same size after region has been divided

The example output can be viewed as Figure 8.

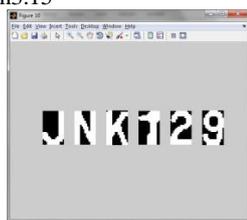

Figure 8. Image output after segmentation process

To wrap up, a good segmentation process should turn out uniform and homogeneous regions with respect to some characteristics such as gray tone or texture as well as simple regions without many small holes. The output from the segmentation will be used in the next stage which is character recognition.

CHARACTER RECOGNITION

Character recognition is the most important task in recognizing the plate number [109]. The recognition of characters has been a problem that has received much attention [110, 111] in the fields of image processing, pattern recognition and artificial intelligence. It is because there is a lot of possibility that the character produced from the normalization step differ from the database. The same characters may differ in sizes, shape and style [110] that could result in recognition of false character, and affect the effectiveness and increase the complexity of the whole system. In Malaysian car plate, there are two groups of character, which is alphabet and numeric. It is important for the system to differentiate the character correctly as sometimes the system may confuse due to the similarities in the form of shape.

When a plate number is put for visual recognition, it is expected to be consisting of one or more characters. However, it may also contain the unwanted information; for example, it may contain pictures and colors that do not provide any useful information to recognize the plate number. Thus, the image is first processed for noise reduction and normalization [111, 112]. Noise reduction is to ensure that the image is free from noise [112]. The normalization is where the isolated characters are resized to fit the characters into a binary window and form the input for recognition process [111, 112]. The characters are segment into a block that contains no extra white spaces in all side of the characters.

Next is the process of digitization [110]. Digitization of an image is converting the individual character into binary matrix based on the specified dimensions. This process will ensure the uniformity of dimensions between the input and stored patterns in the database. For example, in Figure 9, the alphabet A has been digitized into 24x15=360 binary matrix, each having either black or white color pixel [113,114,117,118]. It is important to convert the data into meaningful information. A binary image function can then be assign for each black pixel, the value is 0 (background) and for each white pixel, the value is 1 (foreground) [113, 115].

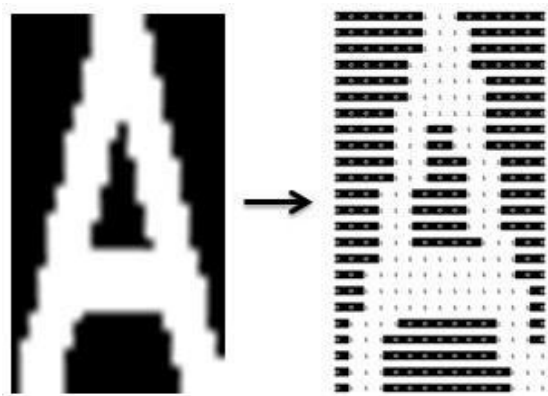

Figure 9. Image digitization

There are a few methods applied for the recognition of characters like template matching, feature extraction, geometric approach, neural network, support vector machine, Hidden Markov Model and Bayes net [111, 112, 116].

A. Template Matching

Template matching is a technique to identify the segmented character [114] by finding the small part in image that match with the template. This method need character image as their template to store in the database [111]. The identification is done by calculating the correlation coefficient where the template the score the highest coefficient is identified as the character of the input [111, 112]. There are three type of matching factor that represent the output which are exact matching, complete mismatching and confused matching [120]. However, due to some similarities in characters, there might be some error during the recognition [121]. Example of character similarities are like, B and 8 or 3, S and 5, Q and G or 0. It should be noted that the size of input image and the template must be exactly the same [114].

B. Neural Network

A Multi-Layer Perceptron Neural Network (MLP NN) [113, 115] in Figure 10 is commonly used in pattern recognition. MLP has been used to solve various problems by training it in supervise learning with the back-propagation algorithm [109, 110, 115, 119]. The training is done in order to make the input leads to a specific target output [111, 116]. The initial weight is randomly generated, and iteratively modified [113]. The weight is modified using the error between the corresponding outputs with desired output. Neural network learns through updating their weight [110]. The purpose of adjusting the weight is to make the output closer to the desired output. It is very important to expand the size of training database in neural network because the efficiency and accuracy of the character recognition will be improved [109].

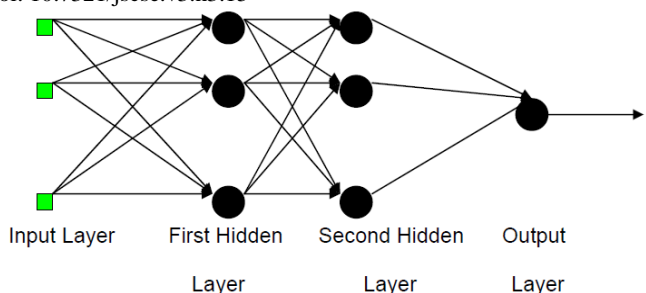

Figure 10. Multilayer perceptron in neural network

### C. Chain Code

Chain code is one of the techniques that are able to do the character recognition process [122]. It is one of the shape representations that are used to represent a boundary of a connected sequence of straight line segments [122]. The representation is based on 4-connectivity and 8-connectivity of the segment that may proceed in clockwise or in anticlockwise direction like in Figure 11 [122, 123, 125]. Collision might be occurred when there is multi-connectivity in the character, and thus multiple chain codes is produces to represent the segment of the character [123, 126]. Besides, the same character might produce different chain code depend on the starting point and their connectivity direction [124]. Thus in order to standardize the character recognition, some additional parameter is needed and calculated [126]. Some constant parameters that need to justify are segment slope angle, character height and the index in the row [126].

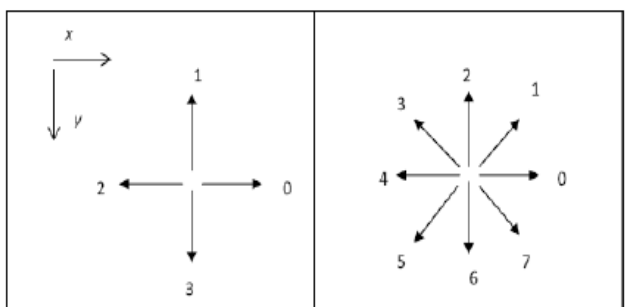

Figure 11. Direction number for 4-directional chain code and 8-directional chain code adapted from N.A. Jusoh, J.M.

### D. Hidden Markov Model

Hidden Markov Model (HMM) is another common used technique for character recognition. HMM is a probabilistic technique [127, 128] that is widely used in pattern recognition area like speech recognition, biological sequence and modeling [129, 130]. A HMM is defined as a doubly stochastic process that is not directly observable (hidden), but can only be observed through another set of stochastic process that produce the sequence of observed character [129, 131]. For character recognition, two main approached is used to construct the model, either for each character or for each word [129]. The advantage of this technique is that it has the ability to learn the similarities and differences between the image samples [127]. For training stage, the sample image must be exactly the same size with the images to deal with [132]. The license plate is represented as a sequence of state as Figure 12, which can generate the observation vector, based on the associated probability distribution. The transition probability is responsible to observe the transition occurred between the states [131, 133]. The parameters or probabilities in HMM are trained using the observation vector extracted from the image samples of license plate [127, 131].

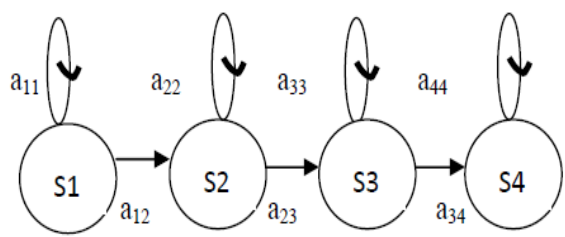

Figure 12. Hidden Markov Model topology for license plate image adapted from S.A. Daramola, E. Adetiba, et. al [127]

The example output can be viewed as Figure 13. The output from the recognition process will be the final output of the license plate recognition in this framework of study.

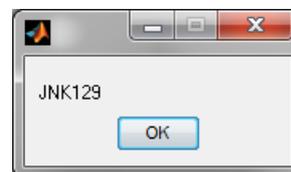

Figure 13. Image output after character recognition process

### VI. CONCLUSION

To conclude this paper, we have presented the review of image processing techniques for license plate recognition with various approaches. The experiment has been done in MATLAB to show the basic process of the image processing especially for license plate in Malaysia case study. There are many more techniques and approaches have been studied for in various stages of image processing as well as there are also lack of studies in image processing stages, for example Pal and Pal in [134] reveals that earlier reviews on colour image segmentation have not given much attention.

### VII. ACKNOWLEDGEMENT

We would like to acknowledge the Malaysian Technical University Network - Centre of Excellence (MTUN CoE) for the funding granted, MTUN/2012/UTeM-FTMK/11 M00019. We also would like to thank all friends and colleagues for their helpful comments and courage.